\documentclass[letterpaper, 10 pt, conference]{ieeeconf}  
\usepackage{url}
\usepackage{hyperref}
\usepackage{adjustbox}

\IEEEoverridecommandlockouts                              

\usepackage{xcolor}
\usepackage{amsmath}
\usepackage{bbm}
\usepackage{graphicx}
\usepackage{duckuments}
\usepackage{capt-of}
\usepackage{cuted}
\usepackage{comment}

\title{\LARGE \bf
Mixture-of-Experts RL for Fault-Tolerant Legged Locomotion
}

\author{Giulio Turrisi$^{1}$, Ozan Pali$^{1}$, Luca Oneto$^{2}$, and Claudio Semini$^{1}$
\thanks{$^{1}$ Dynamic Legged Systems Laboratory, Istituto Italiano di Tecnologia (IIT),
Genova, Italy. E-mail: name.lastname@iit.it}%
\thanks{$^{2}$Dipartimento di Informatica, Bioingegneria, Robotica e Ingegneria dei
Sistemi (DIBRIS), Universita di Genova, Genova, Italy}%
}

\begin{document}

\maketitle
\thispagestyle{empty}
\pagestyle{empty}

\begin{strip}
\vspace{-6em}
\centering

\includegraphics[width=0.325\textwidth]{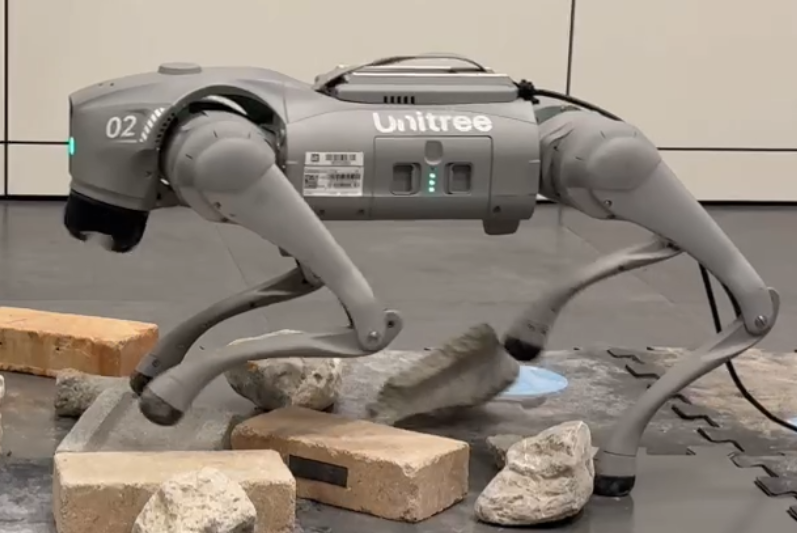}
\includegraphics[width=0.325\textwidth]{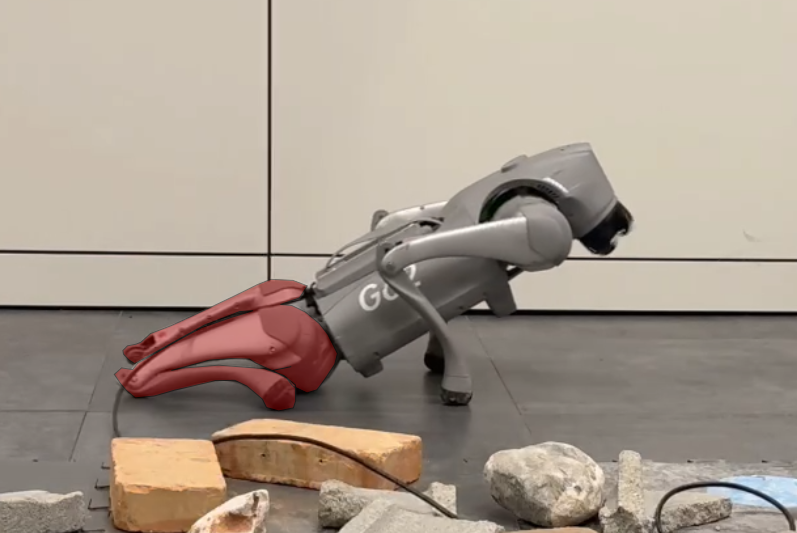}
\includegraphics[width=0.325\textwidth]{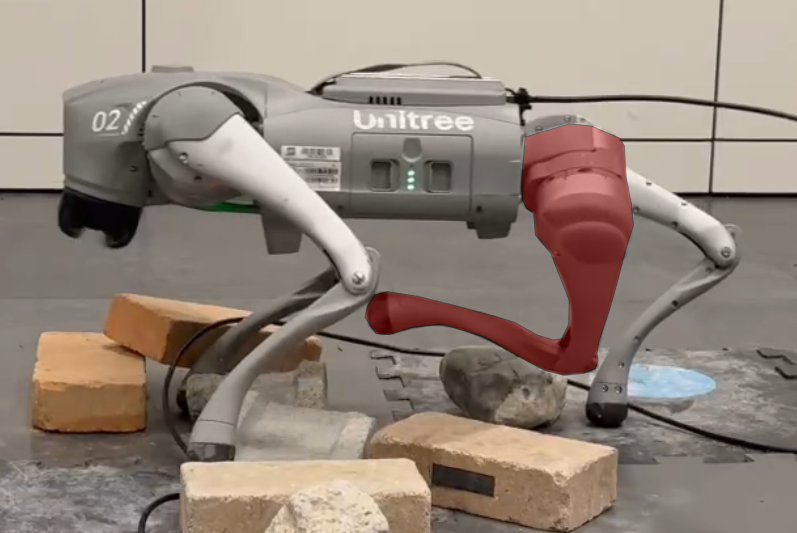}
\vspace{-0.5em}
\captionof{figure}{
Experimental validation of the proposed fault-aware locomotion framework on a Unitree Go2 traversing highly cluttered terrain.
From left to right: locomotion with no leg faults, locomotion with hind-legs actuators failures, and three-legged locomotion. A visual overlay is applied to the faulty legs to highlight the failure condition.
}
\label{fig:teaser}
\vspace{-1.0em}
\end{strip}

\begin{abstract}

Legged robots deployed in planetary exploration and other remote environments must maintain reliable locomotion despite actuator failures and challenging terrain conditions.
Although reinforcement learning has achieved strong results in legged locomotion, monolithic policies can struggle to efficiently represent the diverse control strategies required to compensate for different fault conditions.
In this work, we propose a fault-aware modular control architecture that explicitly leverages fault-diagnosis information to activate specialized control experts associated with distinct actuator failure modes.
Experimental results show that explicit fault-conditioned modular policies consistently outperform monolithic policies of comparable size, achieving higher locomotion performance across failure scenarios.
Moreover, the proposed modular architecture retains competitive performance even under significantly reduced network capacity, highlighting its suitability for compute-constrained robotic platforms, such as those typically employed in space applications.
The code associated with this work is available at: \href{https://github.com/iit-DLSLab/fault-locomotion-isaaclab}
{\url{https://github.com/iit-DLSLab/fault-locomotion-isaaclab}}.

\end{abstract}

\section{INTRODUCTION}

Legged robots are increasingly being considered for planetary exploration \cite{SpaceHopper,ScientificExp}, inspection, and autonomous operation in remote and unstructured environments \cite{vero,BuildingForest}. Their ability to traverse rough terrain, negotiate obstacles, and maintain mobility under challenging conditions makes them attractive candidates for future space missions, particularly when compared with wheeled systems \cite{Kolvenbach2024}. However, long-duration deployment in remote environments exposes robotic platforms to hardware degradation and actuator failures, which can severely compromise locomotion performance. Since physical intervention is often infeasible, such robots must be able to autonomously adapt their locomotion behavior and preserve functionality despite partial hardware failures.

Recent advances in legged locomotion \cite{NonGaited,PerceptiveLoco} have enabled the development of highly capable control policies that achieve robust performance across a wide range of terrains. Among these approaches, reinforcement learning (RL) has demonstrated remarkable robustness and agility \cite{ExtremeParkour,RobotParkour,LearningQuadChallengingTerrain}, paving the way for the safe deployment of legged robots in real-world scenarios. Nevertheless, most existing methods rely on monolithic neural policies that must simultaneously encode terrain perception, locomotion control, and fault adaptation within a single network \cite{luo2023}. As task complexity increases, these competing objectives may require the policy to represent heterogeneous behaviors associated with distinct failure conditions, potentially reducing control performances. 

A common strategy for addressing multimodal control problems is to employ Mixture-of-Experts (MoE) architectures \cite{AdaptiveMixture}, in which multiple specialized policies are coordinated by a learned routing mechanism. In \cite{MoELoco}, the authors demonstrate the benefits of this approach for multimodal locomotion, where a quadruped robot is commanded to execute different gaits, such as trotting, bipedal locomotion, and crawling under obstacles. Similarly, \cite{CMoE} shows the effectiveness of MoE architectures in handling diverse terrain conditions, with each expert specializing in a different locomotion scenario.

\begin{figure*}[h]
\centering
\includegraphics[width=0.99\linewidth]{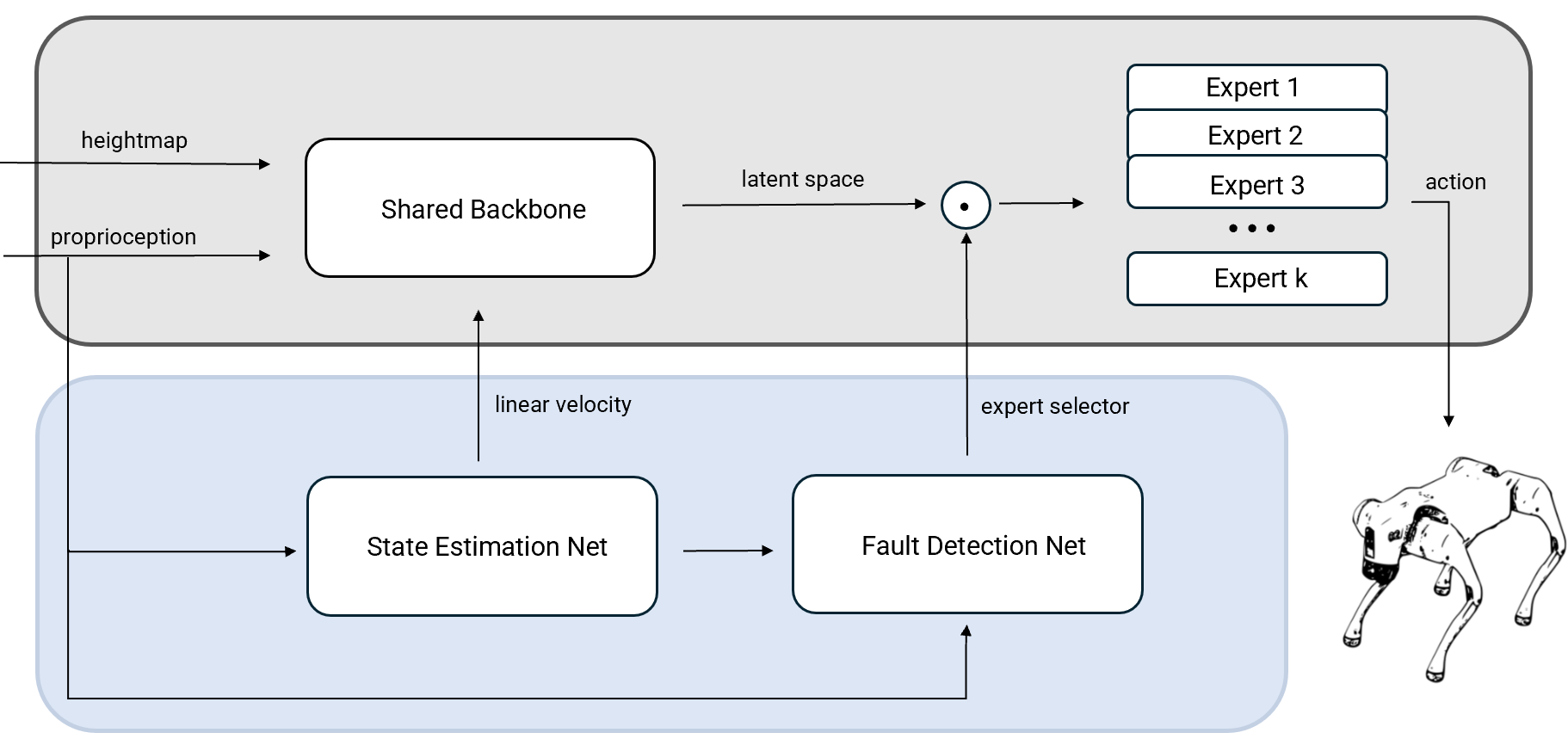}
\caption{Block scheme of the proposed approach. State Estimation and Fault Detection Nets are queried with stacked proprioceptive information, and provide linear velocity and fault diagnosis of the motor's legs. These informations enters to the policy network, which select the appropriate expert to compute the robot's actions.}
\label{fig:block_scheme}
\end{figure*}

While effective, these approaches introduce additional computational overhead and often require the routing network itself to infer the robot's operating mode. In many robotic systems, however, actuator faults can already be identified through onboard health-monitoring systems, diagnostic modules, or past state signals. This information provides an opportunity to directly exploit fault knowledge in the design of the control architecture.

In this work, we propose a fault-aware modular reinforcement learning framework for legged locomotion under actuator failures \ref{fig:teaser}. Instead of relying on learned routing, we explicitly associate specialized control experts with different fault conditions and activate the appropriate expert using estimated fault information. This design enables conditional computation, where only one expert is activated for each fault condition, thereby reducing active inference cost and allowing the controller to specialize recovery behaviors while preserving a compact network structure.

Our results demonstrate that explicit fault-conditioned expert policies consistently win over monolithic Proximal Policy Optimization (PPO) \cite{ppo} baselines across a range of rough-terrain locomotion tasks with actuator failures. Moreover, the proposed architecture maintains competitive performance even when network capacity is reduced, suggesting that structured specialization can provide an effective alternative to simply increasing model size. These properties make the approach particularly attractive for robotic systems operating under onboard computational constraints, such as planetary exploration platforms with limited hardware resources.

\subsection{Contribution}

The main contributions of this work can be summarized as follows:
\begin{itemize}
\item We propose a fault-aware Mixture-of-Experts policy for robust fault recovery in legged locomotion, leveraging explicit expert specialization conditioned on diagnosed actuator failures.
\item We evaluate the effectiveness of the proposed architecture under different network capacities, showing that modular policies can achieve competitive locomotion performance while reducing active computational requirements. This makes them particularly attractive for resource-constrained robotic platforms.
\end{itemize}
Finally, we release the code associated with this work to provide the community with a reproducible baseline for comparison and a foundation for future research.

\subsection{Outline}

The remainder of this paper is organized as follows. Section~\ref{sec:methodology} presents the proposed fault-aware locomotion framework, including the Mixture-of-Experts policy architecture, the fault-detection module, and the state-estimation pipeline adopted for locomotion control. Section~\ref{sec:results} evaluates the proposed approach in simulation and on a real quadruped robot, analyzing its performance under different actuator-failure conditions and network-capacity settings. Finally, Section~\ref{sec:conclusions} discusses the main findings, current limitations, and directions for future work.

\section{Methodology}
\label{sec:methodology}

We address the problem of fault-tolerant locomotion for quadruped robots operating on rough terrain under actuator failures. The proposed framework integrates terrain perception, proprioceptive state estimation, and fault-aware control within a modular reinforcement learning architecture. Unlike conventional monolithic policies, which must represent locomotion behaviors across multiple fault conditions within a single network, our approach explicitly exploits fault information estimated by a dedicated neural module to activate specialized policy experts tailored to specific actuator failure modes.

The overall control pipeline consists of three main components: i) a fault-aware Mixture-of-Experts policy that generates joint-space locomotion commands; ii) a state-estimation module that uses proprioceptive information to estimate the base linear velocity; and iii) a fault-detection module that identifies failed actuators from the robot's recent state history.

A block diagram of the proposed framework is shown in Fig.~\ref{fig:block_scheme}.

\subsection{Fault-Aware Mixture of Experts}
\label{subsec:moe}

Locomotion under actuator failures is characterized by inherently heterogeneous dynamics. Different failure conditions often require distinct compensation strategies, such as redistributing support forces, modifying foot placement, or adopting alternative gait patterns. For instance, a robot operating without failures typically adopts a trotting gait, in which all four legs contribute to generating the ground reaction forces required for locomotion. In contrast, a robot affected by a single-leg failure must redistribute the load among the remaining healthy limbs and adapt its gait accordingly. Under more severe conditions, such as failures involving two legs, locomotion capabilities are significantly degraded, often forcing the robot to rely on dragging motions while its body remains in contact with the ground.

A conventional policy must encode all these behaviors within a single set of parameters. As the number of fault conditions increases, this can lead to representational interference and reduced learning efficiency. To address this issue, we propose a fault-aware Mixture-of-Experts architecture in which specialized experts are associated with different actuator failure conditions. Unlike classical MoE approaches, which learn a routing function, our method leverages fault information provided by the diagnostic subsystem to perform explicit expert selection. This design simplifies the routing process and reduces the computational cost during deployment.

The observations of the locomotion policy, i.e., the actor, are primarily based on proprioceptive information, such as:

\begin{equation}
    \mathbf{o_\mathrm{p}} = (\mathbf{v}, \mathbf{w},
    \bar{\mathbf{g}},
    \mathbf{q}, \dot{\mathbf{q}})
    \label{eq:observation_proprio}
\end{equation}

\noindent where $\mathbf{v}$ and $\boldsymbol{\omega}$ denote the base linear and angular velocities of the robot in the base frame, respectively; $\bar{\mathbf{g}}$ is the projection of the gravity vector in the base frame; while $\mathbf{q}$ and $\dot{\mathbf{q}}$ denote the joint positions and velocities. These quantities are stacked over a temporal horizon $H$ to provide the policy with access to past observations, thereby improving its adaptability and robustness.

Finally, we append to \eqref{eq:observation_proprio} visual terrain information, denoted by $\mathbf{h}$, which encodes the local terrain geometry as a heightmap.

The final policy observation is:

\begin{equation}
    \mathbf{o} = (
    \mathbf{o}^\mathrm{prop},
    \mathbf{o}^\mathrm{terrain},
    \mathbf{v}^\mathrm{des},
    \mathrm{w}^\mathrm{des})
\label{eq:observation_ref}
\end{equation}

where $\mathbf{v}^{\mathrm{des}}$ and $\omega^{\mathrm{des}}$ denote the reference commands to be tracked by the policy. The critic shares the same observation structure as the actor, but is additionally provided with privileged information \cite{pinto2018asymmetric}, such as PD gains, contact states, and swing/stance timings of the robot legs.

Both the actor and the critic follow the same architectural structure: a shared encoder first processes the observation vector $\mathbf{o}$ to extract a compact latent representation

\begin{equation}
\mathbf{o}^\mathrm{lat} = \mathbf{z}(\mathbf{o}).
\end{equation}

which is then provided to a collection of specialized experts,
$
\pi_1,
\pi_2,
\ldots,
\pi_K
$,
each responsible for a specific fault condition.

In standard MoE architectures, each expert is selected or weighted by a gating network $\mathbf{g}$, which computes

\begin{equation}
\mathbf{\hat{g}_i} = \mathrm{softmax}\!\left(\mathbf{g}(\mathbf{\mathbf{o}^\mathrm{lat}})\right)[i]
\label{eq:moe_gating}
\end{equation}

\begin{equation}
\mathbf{a} = \sum_{i=1}^{N} \mathbf{\hat{g}}_\mathrm{i} \cdot   \mathbf{e}_\mathrm{i}(\mathbf{\mathbf{o}^\mathrm{lat}})
\label{eq:moe_action}
\end{equation}

where $\mathbf{e}_i$ denotes the dedicated layers of the $i$-th expert. This formulation is flexible when the appropriate expert cannot be determined a priori, since the gating network learns to orchestrate the contributions of the different experts. In our setting, however, the diagnosed failure mode $k$ is estimated by the fault-detection module, as described in Sec.~\ref{subsec:faults_detection}. Therefore, the corresponding expert can be directly activated through a simpler one-hot selection mechanism.

This design provides an additional advantage beyond fault-specific policy specialization, namely reduced inference complexity. Assuming $K$ experts, the total number of parameters grows linearly with $K$. However, at runtime, only the expert associated with the diagnosed fault condition contributes to the control action. Consequently, the active computational cost is given by

\begin{equation}
C_{\mathrm{active}} =
C_{\mathrm{enc}} + C_k,
\end{equation}

where $C_{\mathrm{enc}}$ denotes the computational cost of the shared encoder, and $C_k$ denotes the computational cost of the selected expert. This contrasts with dense-routing Mixture-of-Experts architectures, in which the output is obtained by combining the predictions of all experts, resulting in the following computational cost:

\begin{equation}
C_{\mathrm{dense}} =
C_{\mathrm{enc}} +
\sum_{k=1}^{K} C_k.
\end{equation}

It should be noted that modern sparse-routing MoE architectures \cite{SparseGate} can achieve a computational complexity similar to that of the proposed approach by activating only a small subset of experts at inference time. However, unlike sparse-routing methods, our framework does not require learning a routing function, since expert selection is directly determined by the fault-diagnosis module.

\begin{table*}[h]
\vspace{0.3cm}
\renewcommand{\arraystretch}{1.05}
\centering
\caption{Locomotion policy rewards. The last column indicates the failure regimes in which each term is active}
\begin{tabular}{c|c|c}
\hline
\textbf{Type} & \textbf{Expression} & \textbf{Active} \\ 
\hline
base linear velocity 
& $\exp(-\|\mathbf{v}^{\mathrm{des}}-\mathbf{v}\|^2/0.05)$ 
& all $n_f$ \\

yaw rate 
& $\exp(-(w_z^{\mathrm{des}}-w_z)^2/0.25)$ 
& all $n_f$ \\

base height 
& $\exp(-(h^{\mathrm{des}}_{\mathrm{base}}+h^{\mathrm{terrain}}-h_{\mathrm{base}})^2/0.01)$ 
& $n_f<2$ \\

base pitch orientation 
& $-(\theta^{\mathrm{terrain}}-\theta)^2$ 
& $n_f\leq1$ \\

base roll orientation 
& $-(\phi^{\mathrm{terrain}}-\phi)^2$ 
& all $n_f$ \\

base vertical velocity 
& $-v_z^2$ 
& $n_f\neq2$ \\

front hip height 
& $\exp(-\sum_{i=1}^{4}(h^{\mathrm{des}}_{\mathrm{hip}}+h^{\mathrm{terrain}}-h_{\mathrm{hip}})^2/0.01)$ 
& $n_f=2$ \\

joints torque 
& $-\|\boldsymbol{\tau}\odot\mathbf{f}\|^2$ 
& active joints \\

joints acceleration 
& $-\|\ddot{\mathbf{q}}\odot\mathbf{f}\|^2$ 
& active joints \\

joints power 
& $-|\dot{\mathbf{q}}\odot\boldsymbol{\tau}\odot\mathbf{f}|$ 
& active joints \\

action rate 
& $-\|(\mathbf{a}_k-\mathbf{a}_{k-1})\odot\mathbf{f}\|^2$ 
& active joints \\

action smoothness 
& $-\|(\mathbf{a}_k-2\mathbf{a}_{k-1}+\mathbf{a}_{k-2})\odot\mathbf{f}\|^2$ 
& active joints \\

undesired contact 
& $-\sum_i \mathbbm{1}_{\mathrm{contact},i}$ 
& trunk: $n_f<2$; legs: healthy legs \\

stance contact suggestion 
& $\frac{1}{4}\sum_{i=1}^{4}c_i l_i$ 
& all $n_f$ \\

feet air time 
& $\sum_{i=1}^{4} t^{\mathrm{air}}_i l_i$ 
& healthy legs \\

feet air-time variance 
& $-\mathrm{Var}(t^{\mathrm{air}},t^{\mathrm{contact}})$ 
& healthy legs \\

feet slide 
& $-\sum_{i=1}^{4}\|\mathbf{v}_{\mathrm{foot},i}^{xy}\|c_i l_i$ 
& healthy legs \\

feet height clearance 
& $\sum_{i=1}^{4}\exp(-(\mathbf{p}_\mathrm{foot}^{\mathrm{des, z}}+h^{\mathrm{terrain}}-\mathbf{p}_\mathrm{foot}^{\mathrm{z}})^2/0.01))$ 
& all $n_f$ \\

feet to hips position 
& $-\sum_{i=1}^{4}\|\tilde{\mathbf{p}}_{\mathrm{foot},i}^{xy}
-\tilde{\mathbf{p}}_{\mathrm{hip},i}^{xy}\|l_i$ 
& all $n_f$ \\

CoM support polygon 
& $\mathbbm{1}(\mathbf{p}_{\mathrm{CoM}}\in\mathcal{P}_{\mathrm{support}})$ 
& $n_f=1$ \\

feet vertical surface contact 
& $-\mathbbm{1}(\|\mathbf{f}_{xy}\|>4|f_z|)$ 
& healthy legs \\
\hline
\end{tabular}
\label{tab:rewards}
\end{table*}

\subsection{State Estimation}
Model-based state-estimation techniques have proven effective under nominal operating conditions \cite{muse}, since they rely on assumptions about contact configurations and robot dynamics. However, these assumptions may no longer hold in the presence of severe actuator failures. For instance, under double-leg failures, the robot may partially drag its body on the ground, producing contact patterns that differ substantially from those observed during nominal locomotion. As a result, the performance of contact-based estimators and model-based observers can degrade significantly.

\begin{figure*}[t]
\centering
\includegraphics[width=0.245\linewidth]{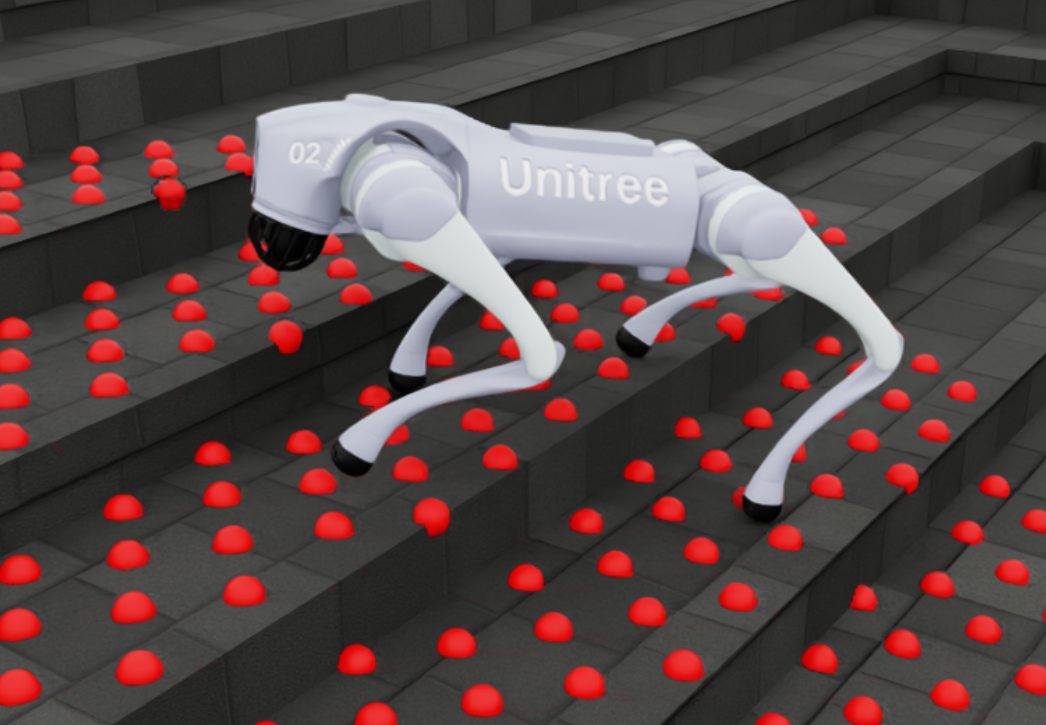}
\includegraphics[width=0.245\linewidth]{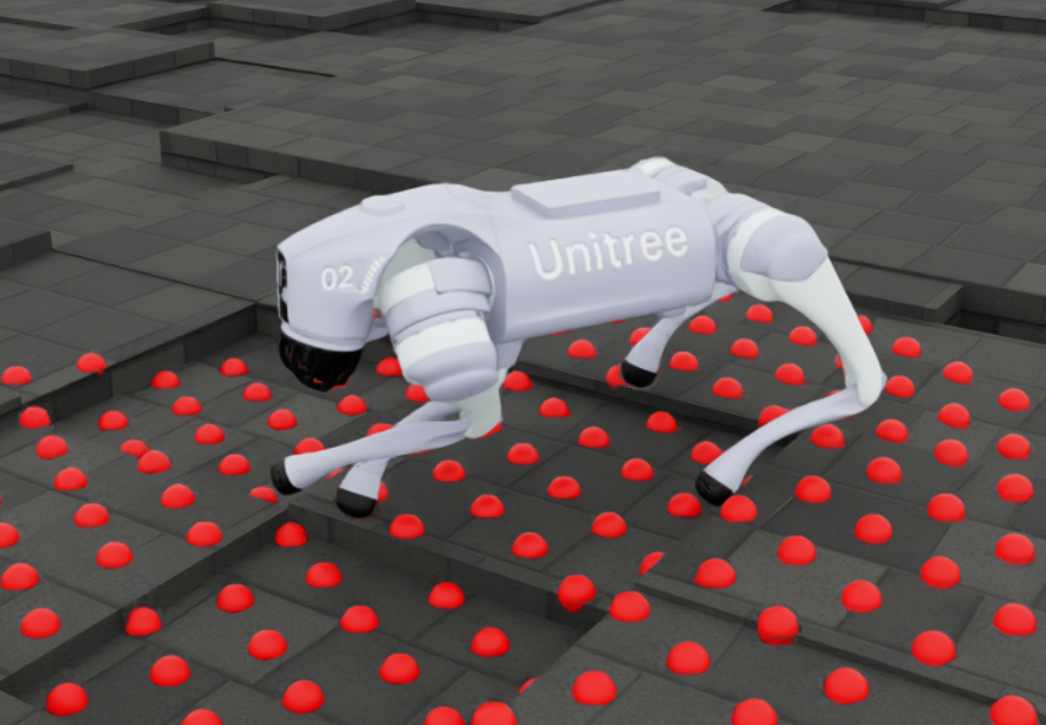}
\includegraphics[width=0.245\linewidth]{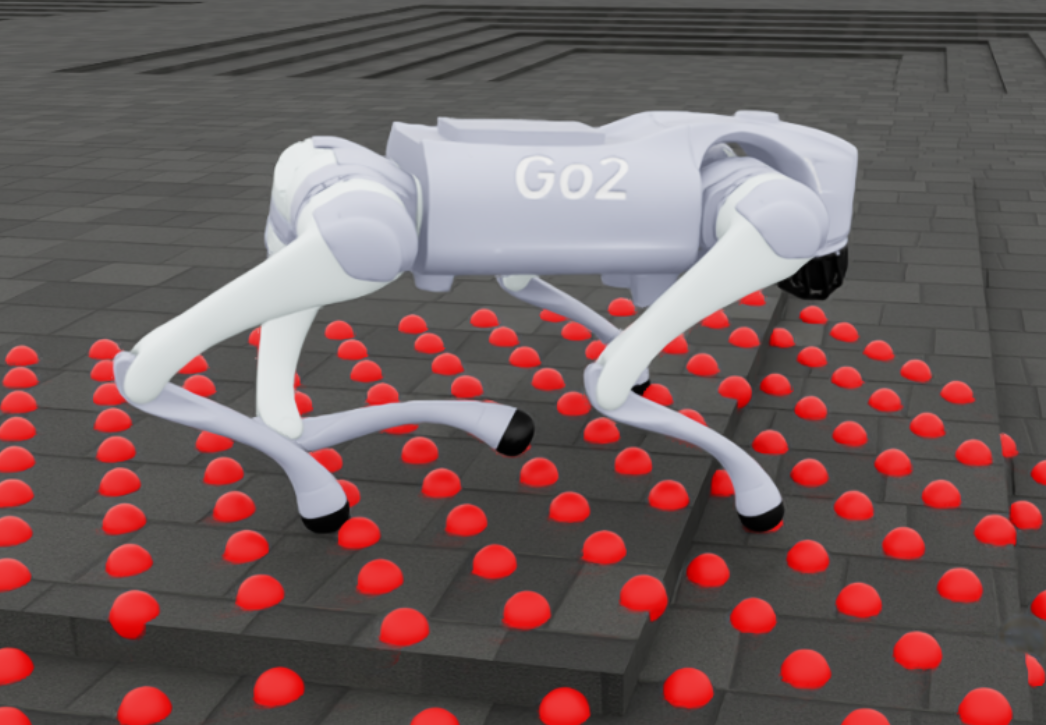}
\includegraphics[width=0.245\linewidth]{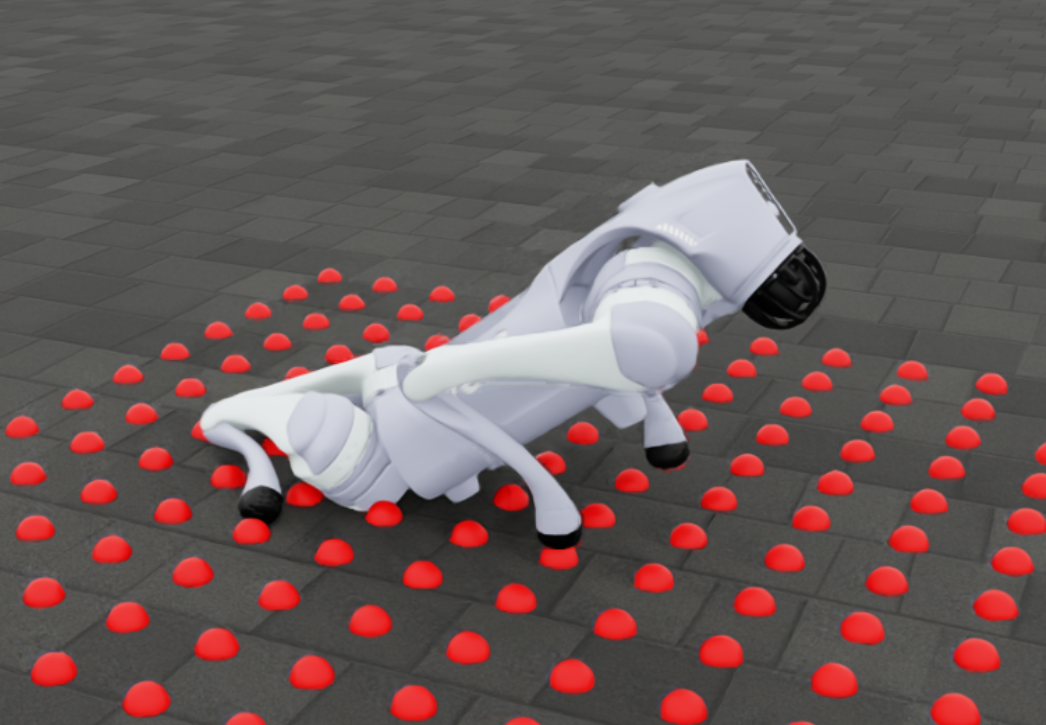}
\caption{
The four different fault modalities. From left to right: \textit{No Fail}, \textit{LF Fail}, \textit{RH Fail}, and \textit{LH+RH Fail}. Red dots represent the heightmap that enters as an observation in the policy network.}
\label{fig:sim}
\end{figure*}

To address these limitations, we adopt a concurrent learning-based state-estimation approach \cite{concurrentse}. Following this formulation, we train a neural network to estimate the base linear velocity directly from sensors observations. The estimator receives

\begin{equation} 
\hat{\mathbf{v}} = f_{\mathrm{est}}(\mathbf{\bar{o}}) 
\end{equation} 

where $\mathbf{\bar{o}}$ is defined similar as in~\eqref{eq:observation_ref}, but with the replacement of the linear velocity with the IMU linear acceleration. This formulation avoids the need for explicit contact modeling and naturally accommodates changes in the robot dynamics induced by actuator failures.

The network is trained in simulation using privileged ground-truth measurements of the robot base velocity. Given the predicted velocity $\hat{\mathbf{v}}$ and the ground-truth velocity $\mathbf{v}$, the estimator is optimized using a mean-squared-error objective:

\begin{equation}
    \mathcal{L}_{\mathrm{est}} =
    \left\|
    \mathbf{v} - \hat{\mathbf{v}}
    \right\|_2^2.
    \label{eq:estimator_loss}
\end{equation}

The resulting velocity estimate is then provided to the locomotion policy.

\subsection{Faults Detection}
\label{subsec:faults_detection}
A key assumption of the proposed framework is the availability of information about the current actuator failure condition. While this information can be directly accessed in simulation, real robotic systems may require dedicated mechanisms to identify hardware degradations and actuator malfunctions during operation. Following the approach of RMA~\cite{KumarA-RSS-21}, we formulate the actuator fault identification process as a supervised learning problem. More specifically, we train a neural network $f_{\mathrm{fault}}$ that receives the same observation history of the state estimation network, and predicts the current fault condition:

\begin{equation} 
\hat{\mathbf{f}} = f_\mathrm{fault}(\mathbf{\bar{o}}) 
\end{equation} 

The use of a temporal observation window is particularly important, since actuator failures cannot always be reliably identified from a single state measurement. 

The output $\hat{\mathbf{f}}$ represents a categorical distribution over the set of considered fault conditions and is used to determine which locomotion expert should be activated. During training, supervision is obtained directly from the fault labels available in simulation, and the network is optimized using a standard cross-entropy loss:

\begin{equation}
\mathcal{L}_{\mathrm{fault}}
=
-\frac{1}{N_f}
\sum_{i=1}^{N_f}
\left[
f_i \log(\hat{f}_i)
+
(1-f_i)\log(1-\hat{f}_i)
\right]
\end{equation}

where $N_f$ denotes the number of fault classes, and $f_i$ denotes the ground-truth fault label. The resulting architecture enables the system to automatically identify actuator failures directly from proprioceptive measurements and is trained concurrently with the locomotion policy.

\section{Results}
\label{sec:results}

The training and evaluation were performed using the IsaacLab framework \cite{mittal2025isaaclab} on the Unitree Go2 robot\footnote{\url{https://www.unitree.com/go2}}. The locomotion policy was trained using the reward function reported in Table~\ref{tab:rewards}. The reward combines task-oriented objectives, such as velocity tracking and terrain-aware pose adaptation, where $h$ denotes the base height and, $\theta$, $\phi$ the base pitch and roll respectively, with regularization terms that promote smooth and energy-efficient motions. These include joint energy, computed from the product of the joint torques $\boldsymbol{\tau}$ and joint velocities $\dot{\mathbf{q}}$; joint-acceleration penalization through $\ddot{\mathbf{q}}$; feet air time $\mathbf{t}^{\mathrm{air}}$, which penalizes excessively long swing phases; feet air-time variance, which encourages similar contact and swing timings across all legs; and stance-contact suggestion, which uses the IsaacLab contact sensors $\mathbf{c}$ on the robot feet to encourage foot contact when the desired velocity is zero.

Fault-specific reward terms are activated depending on the number of failed legs, denoted by $n_f$. A leg $l_i$ is considered failed whenever at least one of its actuated joints experiences a fault. For example, body-height and orientation objectives are emphasized when sufficient support contacts are available, while they are disabled in the case of two-leg failures, i.e., when $n_f = 2$. In the case of one single leg failure instead, the policy is encouraged to shift the center of mass $\mathbf{p}_{\mathrm{CoM}}$ toward the support region defined by the remaining healthy legs, thereby improving locomotion robustness. This formulation allows the policy to progressively adapt its behavior as the severity of the failure increases. Table~\ref{tab:rewards} summarizes all reward terms and the fault regimes in which they are active.

To evaluate the proposed fault-aware Mixture-of-Experts architecture, we focus on four representative failure conditions, as shown in Fig.~\ref{fig:sim}:
\begin{itemize}
\item \textit{No Fail}: nominal locomotion with all actuators operational;
\item \textit{LF Fail}: failure of the left-front leg;
\item \textit{LH Fail}: failure of the left-hind leg;
\item \textit{LH+RH Fail}: simultaneous failure of both hind legs.
\end{itemize}

These scenarios span increasing levels of locomotion degradation, ranging from nominal trotting to severely impaired locomotion requiring body-dragging behaviors. Furthermore, they capture the main classes of actuator failures while avoiding redundant evaluations arising from the inherent symmetries of quadrupedal locomotion. Under standard bilateral symmetry, failures affecting the right-front leg are expected to exhibit behaviors analogous to left-front failures, while right-hind failures are symmetric to left-hind failures \cite{ordonez2025morphosymm}. Consequently, the selected fault conditions provide a representative subset of the full fault space while significantly reducing the experimental burden.

For each failure condition, we compare the proposed fault-aware expert policy against a monolithic PPO baseline. The comparison focuses on two aspects:
\begin{enumerate}
\item locomotion performance, measured through the cumulative episodic reward;
\item learning performance degradation, analyzed by studying how network size affects the quality of the final policy.
\end{enumerate}

The following two subsections report the quantitative results obtained for the considered fault scenarios in simulation and on real hardware.

\subsection{Simulations}

\begin{figure}[t]
\centering
\includegraphics[width=0.99\linewidth]{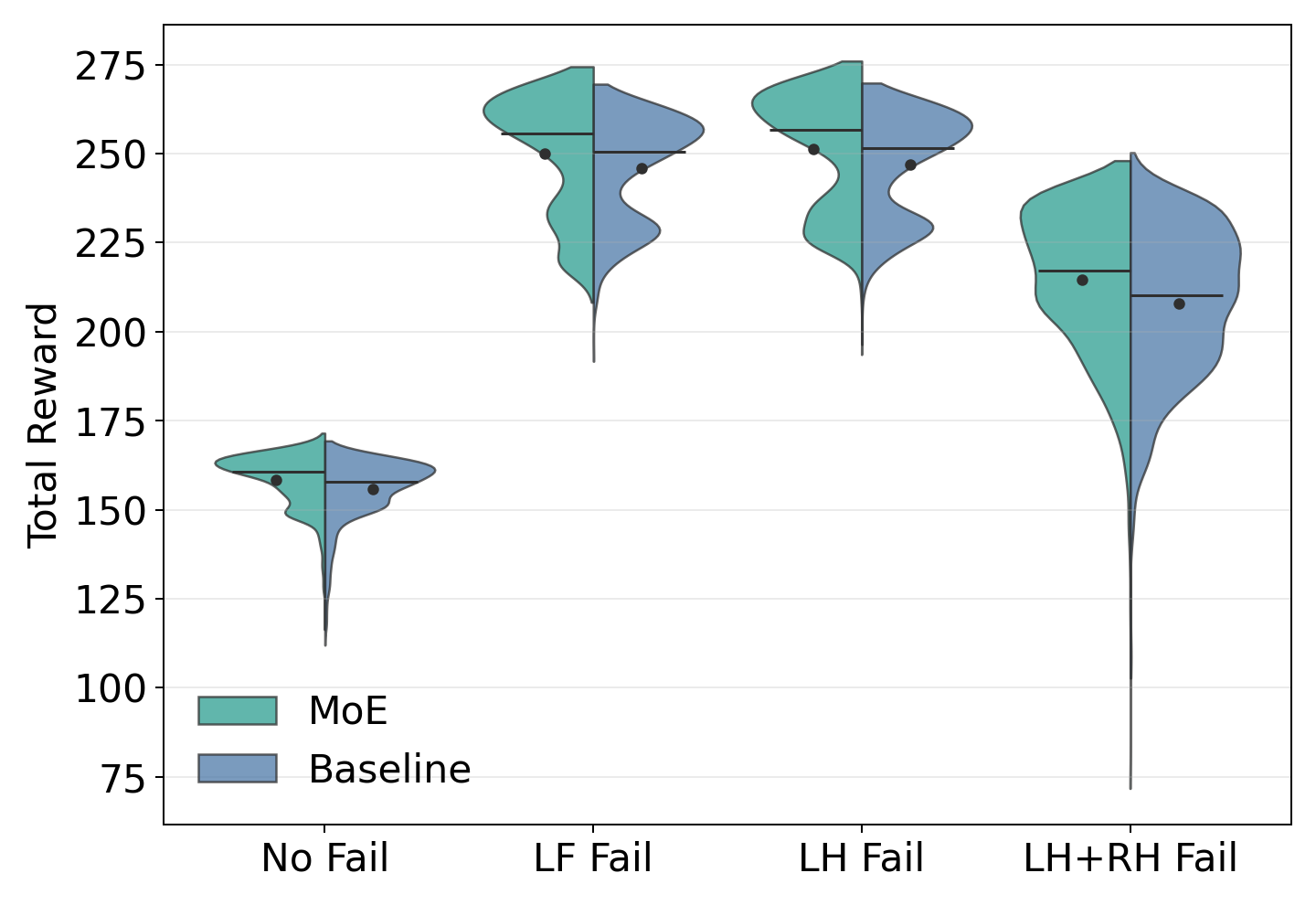}
\caption{Violin plots showing the distribution of rewards obtained across $8,192$ robots, comparing our method (MoE) with the monolithic baseline under four different fault conditions.}
\label{fig:large}
\end{figure}

\begin{figure}[h]
\centering
\includegraphics[width=0.95\linewidth]{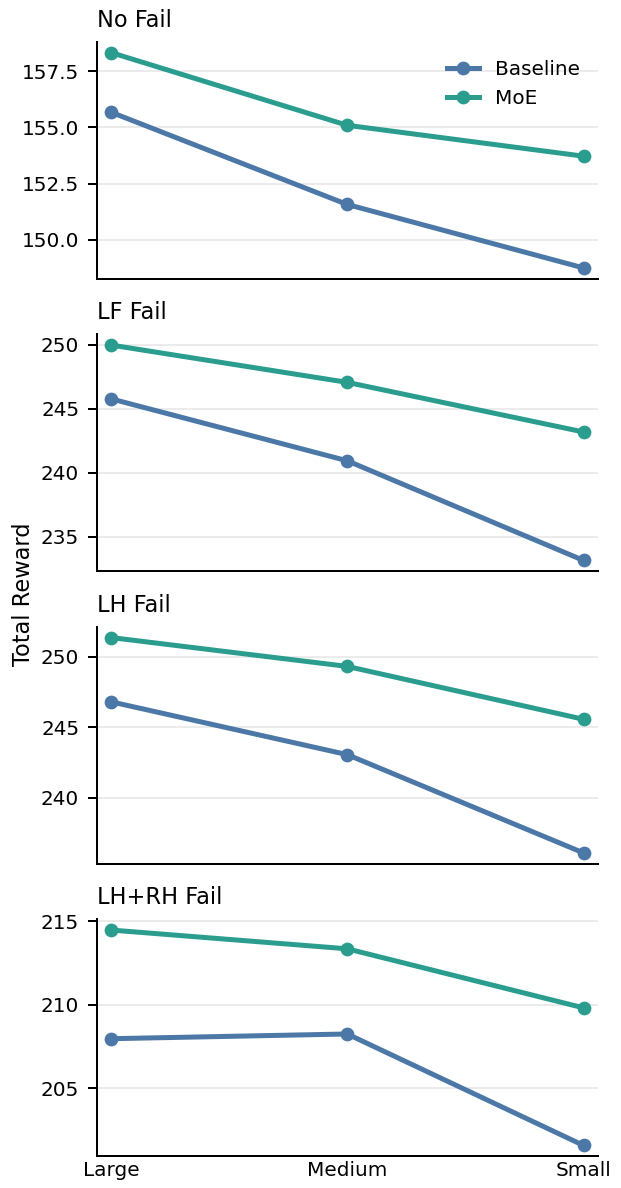}
\caption{Comparison between the proposed Mixture-of-Experts policy (MoE) and a monolithic PPO baseline across the considered fault scenarios. The actor network capacity is progressively reduced from Large to Small, while the critic architecture remains unchanged.}
\label{fig:comparison}
\end{figure}

Figure~\ref{fig:large} reports the distribution of cumulative episodic rewards obtained under the four considered fault conditions, while Fig.~\ref{fig:comparison} analyzes the effect of reducing the policy network capacity.

Figure~\ref{fig:large} shows that the proposed architecture consistently achieves higher performance than the monolithic baseline across all considered fault scenarios. In the nominal condition (\textit{No Fail}), both approaches achieve comparable locomotion performance, indicating that expert specialization does not compromise the ability of the robot to perform standard locomotion. However, as the severity of the fault increases, the advantage of the proposed architecture becomes progressively more evident.

In the single-leg failure scenarios (\textit{LF Fail} and \textit{LH Fail}), the Mixture-of-Experts policy achieves higher cumulative rewards and exhibits a more concentrated reward distribution. This suggests that explicit fault-conditioned specialization enables the controller to learn more effective compensation strategies, resulting in improved locomotion robustness and consistency. The largest performance improvement is observed in the \textit{LH+RH Fail} condition, where both hind legs are disabled and locomotion becomes significantly more challenging. In this severely degraded regime, the robot is often forced to rely on dragging-like behaviors, which differ substantially from standard trotting locomotion. The proposed architecture maintains substantially higher rewards than the monolithic policy, demonstrating the benefit of associating dedicated experts with specific fault conditions when the underlying locomotion strategies become highly heterogeneous.

Figure~\ref{fig:comparison} evaluates the sensitivity of the proposed approach to network capacity. Three policy architectures are considered, denoted as \textit{Large}, \textit{Medium}, and \textit{Small}. All policies are implemented as standard multilayer perceptrons composed of three hidden layers with dimensions $[128,128,128]$, $[64,64,64]$, and $[32,32,32]$, respectively. To isolate the effect of reducing the actor capacity, the critic architecture is kept unchanged across all experiments. Specifically, all critic networks employ three hidden layers of size $[128,128,128]$, matching the architecture used in the \textit{Large} configuration.

As expected, reducing the network size leads to a degradation in locomotion performance for both approaches. However, the proposed Mixture-of-Experts architecture appears to be less affected by this reduction than the monolithic PPO baseline, with its \textit{Small} variant achieving performance levels comparable to those of the \textit{Large} monolithic baseline. This behavior is consistently observed across all fault conditions and becomes particularly evident in the \textit{LH+RH Fail} scenario.

These results indicate that the proposed architecture is capable of preserving fault-specific locomotion skills even when the capacity of the individual networks is reduced. In contrast, the monolithic policy must simultaneously encode nominal locomotion and fault-recovery behaviors within a single set of parameters, making it more sensitive to reductions in model capacity. The MoE architecture can instead exploit a larger overall parameterization through its expert structure, as discussed in Sec.~\ref{subsec:moe}.


\subsection{Experiments}
We conducted experiments on a real Unitree Go2 robot to demonstrate the applicability of the proposed approach in real-world scenarios (see Fig. \ref{fig:teaser}). The robot was first commanded via joystick to traverse a rocky terrain (\textit{No Fail} scenario). Subsequently, both hind legs were deactivated (\textit{LH+RH Fail} scenario), and the robot was commanded to locomote on flat ground using only its forelegs to reach again the initial starting point. We then reactivated all the leg, and then finally induced a left-hind leg failure (\textit{LH Fail} scenario), resulting in a three-legged configuration. In this setting, the robot was again commanded to traverse the same rocky terrain. The reader is referred to the supplementary video for additional details.

\section{Conclusions and Limitations}
\label{sec:conclusions}

In this work, we presented a fault-aware Mixture-of-Experts reinforcement learning framework for fault-tolerant quadrupedal locomotion. The proposed architecture explicitly exploits fault-diagnosis information to activate specialized experts associated with different actuator failure conditions, thereby avoiding the need for a learned routing mechanism while enabling conditional computation.

Our evaluation showed that the proposed approach outperforms a monolithic PPO baseline. The performance gains become increasingly evident as the severity of the failure increases, suggesting that expert specialization is particularly beneficial when the underlying locomotion strategies become highly heterogeneous. Furthermore, the proposed architecture maintains competitive locomotion performance even when the network capacity is significantly reduced, highlighting its potential for space robotics applications, where onboard computational resources are limited.

Despite these promising results, an important limitation of the proposed approach was observed during training. Since the expert networks do not share all parameters, the resulting policy-gradient estimates are noisier than those of a monolithic baseline. Intuitively, each expert is updated only using trajectories generated by the subset of agents associated with its corresponding fault condition, effectively reducing the number of samples available for policy optimization.

To mitigate this issue, all experiments were conducted using more than $16{,}384$ parallel simulation environments, resulting in approximately $4{,}096$ agents per fault condition. While this was sufficient for the fault scenarios considered in this work, the sample efficiency of the proposed approach is expected to decrease as the number of experts increases. Consequently, scaling the architecture to a larger number of fault conditions may require either additional simulation environments or more sophisticated parameter-sharing strategies among experts.


\bibliographystyle{IEEEtran}
\bibliography{biblio}

\end{document}